\documentclass[11pt, a4paper]{deepmind}
\usepackage{graphicx}

\usepackage[utf8]{inputenc} 
\usepackage[T1]{fontenc}    
\usepackage{url}            
\usepackage{booktabs}       
\usepackage{amsfonts}       
\usepackage{nicefrac}       
\usepackage{microtype}      
\usepackage[dvipsnames]{xcolor}
\usepackage{wrapfig}
\usepackage{algorithm}
\usepackage{algpseudocode}

\usepackage{caption}
\usepackage{array}
\newcolumntype{P}[1]{>{\centering\arraybackslash}p{#1}}

\usepackage{amsmath}
\DeclareMathOperator*{\argmax}{argmax}

\usepackage[sorting=none]{biblatex} 
\usepackage{hyperref}

\title{Schema-learning and rebinding as mechanisms of in-context learning and emergence}

\renewcommand{\today}{}

\author[1]{Sivaramakrishnan Swaminathan}
\author[1]{Antoine Dedieu}
\author[1]{Rajkumar Vasudeva Raju}
\author[1]{Murray Shanahan}
\author[1]{Miguel L\'azaro-Gredilla}
\author[1]{Dileep George}
\affil[1]{Google DeepMind}

\correspondingauthor{sivark@google.com}

\begin{abstract}
In-context learning (ICL) is one of the most powerful and most unexpected capabilities to emerge in recent transformer-based large language models (LLMs). Yet the mechanisms that underlie it are poorly understood. In this paper, we demonstrate that comparable ICL capabilities can be acquired by an alternative sequence prediction learning method using clone-structured causal graphs (CSCGs). Moreover, a key property of CSCGs is that, unlike transformer-based LLMs, they are {\em interpretable}, which considerably simplifies the task of explaining how ICL works. Specifically, we show that it uses a combination of (a) learning template (schema) circuits for pattern completion, (b) retrieving relevant templates in a context-sensitive manner, and (c) rebinding of novel tokens to appropriate slots in the templates. We go on to marshall evidence for the hypothesis that similar mechanisms underlie ICL in LLMs. For example, we find that, with CSCGs as with LLMs, different capabilities emerge at different levels of overparameterization, suggesting that overparameterization helps in learning more complex template (schema) circuits. By showing how ICL can be achieved with small models and datasets, we open up a path to novel architectures, and take a vital step towards a more general understanding of the mechanics behind this important capability.
\end{abstract}

\begin{document}

\maketitle

\section{Introduction}

In a pre-trained sequence model, {\em in-context learning} (ICL), or {\em few-shot prompting}, is the ability to learn a new task from a small set of examples presented within the context (the prompt) at inference time. Surprisingly, large language models (LLMs) trained on sufficient data exhibit ICL, even though they are trained only with the objective of next token prediction \cite{brown2020language,webb2022emergent}. A good deal of the ongoing excitement surrounding LLMs arises from this unexpected capacity, since it dramatically enlarges their set of potential applications. Attempts to understand this capability are ongoing and take a variety of forms, including higher-level normative accounts using Bayesian inference \cite{xie2021explanation}, and mechanistic explanations involving implicit gradient descent \cite{dai2022can} or induction heads \cite{olsson2022context}. Despite this, the mechanisms that underlie ICL in LLMs remain somewhat mysterious.

In this paper, we take an alternative approach. We reveal the conditions that drive ICL in a different sequence learning model called a clone-structured causal graph (CSCG) \cite{george2021clone, dedieu2019learning}. Using a combination of new and standard datasets, we show how a CSCG assigns non-zero probabilities to sequences never seen during training in a way that, thanks to the model's causal graph structure, is open to explicit and mechanistic interpretation. We hypothesize that similar mechanisms will exist in transformer-based LLMs, and show how this could be the case.

Specifically, we show that ICL in CSCGs can be explained as a combination of (a) learning template circuits for pattern completion, (b) retrieving relevant templates in a context-sensitive manner, and (c) rebinding of novel tokens to appropriate slots in templates \cite{shanahan2022abstraction}. Unlike n-gram models, CSCGs allow transitive generalization in the latent space, which assigns non-zero probabilities to sequences never seen during training in a semantically sensible way, ensuring that the contexts (prompts) used for retrieval are not pure memorizations. In addition, the binding of novel tokens to slots in learned templates allows the same structural knowledge to be applied to entirely novel inputs. By elucidating the principles that underpin the mechanics of ICL, we hope to pave the way for the design of novel architectures for abstraction and generalization, while the building blocks we identify guide the search for mechanistically interpretable \cite{nanda2023progress} and editable \cite{meng2022locating} circuits in transformers \cite{vaswani2017attention}.

\section{Rebinding algorithm for clone-structured causal graphs}

\subsection{Background on clone-structured causal graphs (CSCGs)}
Consider an agent executing a series of discrete actions $a_1,\ldots,a_{N-1}$ with $a_n\in\{1,\ldots, N_\text{actions}\}$, e.g. walking in a room. As a result of each action, the agent receives a perceptually aliased observation \cite{chrisman1992reinforcement}, resulting in the stream of random variables $X_1,\ldots,X_N$ with observed values $x_1,\ldots,x_N$, where each $x_n\in\{1,\ldots, N_\text{obs}\}$. CSCG \cite{george2021clone} is a probabilistic sequence learning model that learns latent graphs to model these action-conditional sequences. A CSCG can recover a graph that represents the latent causal structure \cite{pearl2009causality} of the environment (e.g. the room), which can then be used to plan actions in that environment. CSCGs introduce a latent explanatory variable $Z_n$ at each timestep $n$, with values $z_n\in\{1,\ldots, N_\text{latent}\}$, to disambiguate the perceptually aliased observations. It then models the stream of observations as
\begin{equation*}
P(x_1, \ldots, x_{N} | a_1,\ldots, a_{N-1}) = \sum_{z_1, \ldots, z_n}P(x_1 | z_1) P(z_1)\prod_{n=2}^N P(x_n | z_n) P(z_n| z_{n-1}, a_{n-1}).
\label{eq:cscg}
\end{equation*}

The action-conditional dynamics are represented by a transition tensor $T: ~T_{ijk}= P(Z_n=k| Z_{n-1}=j, a_{n-1}=i)~\forall n$, and the observation probabilities by an emission matrix  $E: ~E_{ij}= P(X_n=j| Z_n=i)~\forall n$. The transition tensor $T$ defines a directed multigraph, whose nodes correspond to the values of $z$. Conditioned on an action, each entry of $T$ is the weight of a directed edge between two nodes (from the row index to the column index of that entry). 
See Fig.~\ref{fig:CSCG-concepts}D for an example of a recovered CSCG latent graph.

CSCG has a deterministic observation model. That is, for each row of $E$, one entry is set to $1$ and the remaining to $0$. Hence, for any latent value $z$, the same observation $x$ is always emitted. Multiple values of $z$ can result in the same observed $x$, making the model overcomplete \cite{sharan2017learning}. Using these latent states, a CSCG can disambiguate multiple aliased percepts (same observation $x$) into distinct causes (different latent values $z$). The restriction of $E$ being deterministic makes CSCGs less general than a hidden Markov model (HMM), but easier to learn \cite{george2021clone}. CSCG can also be used as a language model if the observations correspond to word tokens, with the single action that accesses the next token\footnote{This can be generalized to include actions that skip over one or more next tokens.}.

\subsection{Rebinding in CSCGs}\label{sec:alg_rebinding}

\begin{figure}[t!]
    \centering
    \includegraphics[width=0.95\textwidth]{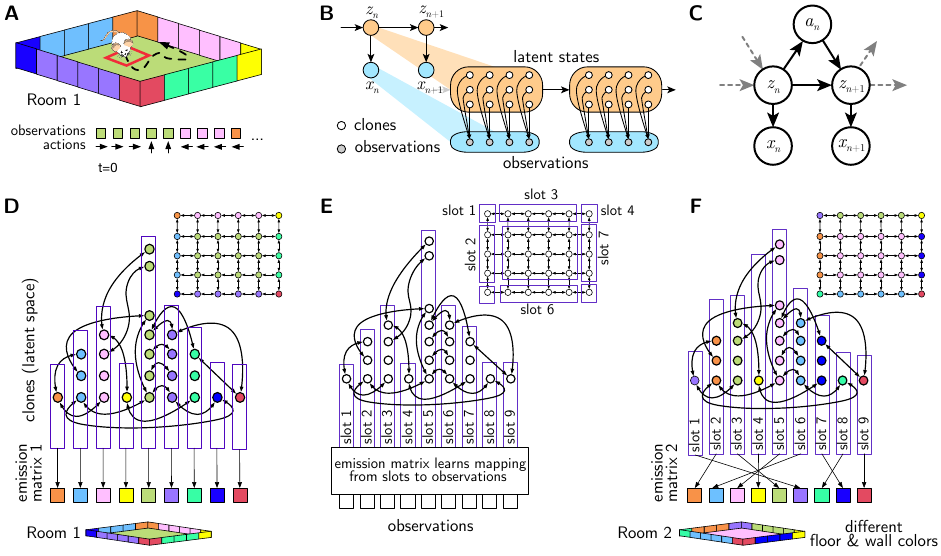}
    \caption{
    {\bf A}. Inducing the structure of the room (\emph{cognitive maps}) from sequential sensory observations is challenging because of perceptual aliasing -- local observations do not identify locations uniquely. {\bf B}. Cloned hidden Markov models (HMMs) \cite{dedieu2019learning}. Each observation is mapped to multiple clone states in the latent space. {\bf C}. Graphical model for CSCGs \cite{george2021clone}, extending cloned HMMs by incorporating actions. CSCGs utilize the latent space to overcome the perceptual aliasing problem. Different clones learn to represent different temporal contexts to recover the latent structure of the room.
    {\bf D}. Learned CSCG for the room shown in panel A consists of a latent transition matrix and an emission matrix. We visualize the model in two ways: (i) stacking clone states for respective observations into columns, and (ii) clones as nodes in a transition graph, colored with their respective emissions.
    {\bf E}. The emission matrix imposes a \emph{slot} structure -- nodes within the same slot are constrained to bind to the same observation. A new environment with the same latent structure but different observation mapping (Room2, for example) can be learned quickly by freezing the transition graph and the slot structure, and learning a new emission matrix by rebinding slots to a new set of observations. {\bf F}. CSCG for a \emph{different room} learned purely through rebinding.}
    \label{fig:CSCG-concepts}
\end{figure}

When an agent encounters a new environment with a similar structure, but different observations, it can learn that environment faster by reusing the latent graph $T$ (what we call a \emph{schema}) \cite{guntupalli2023graph} from prior experience and relearning just the emission matrix. We call this process \emph{rebinding}. Rebinding can be interpreted as a soft intervention on the agent's model \cite{peters2017elements, eaton2007exact}. See Fig.~\ref{fig:CSCG-concepts}D~\&~F for examples of two rooms that share the same latent structure but different observations. 
When a new emission matrix binds to an existing schema, it has to respect the \emph{clone structure} of the original emission matrix (Fig.~\ref{fig:CSCG-concepts}E). 
The clone structure function ${\cal C}(\cdot)\in 1,\ldots,N_{\text{obs}}$ partitions the latent state in $N_{\text{obs}}$ slots: 
two latent states \(z=i\) and \(z=i'\) belong to the same clone slot iff \({\cal C}(i) = {\cal C}(i')\). An emission matrix respects the clone structure ${\cal C}$ if the condition \({\cal C}(i)={\cal C}(i') \implies E_{ij}=E_{i'j}~\forall~i,i',j\) is satisfied. 
The 3-tuple $\{T, {\cal C}, E\}$ defines a \emph{grounded schema}, the tuple $\{T, {\cal C}\}$ defines an \emph{ungrounded schema with clone structure} and $T$ alone is a \emph{schema} \cite{guntupalli2023graph}.

\subsubsection{Fast rebinding by attending to surprise}
Often, environment changes are localized such that most of the latent structure and observation mapping is preserved while just a few observations need to be rebound: for example, just replacing the carpet in a room while the wall colors remain the same, or getting exposed to a new word in a familiar context. This insight can be utilized to derive an algorithm that focuses the update of the emission matrix only to those observations that were found surprising by the existing model.

Suppose that at test time, a grounded schema $\{T, {\cal C}, E^0\}$ is exposed to a sequence with novel observations. Algorithm~\ref{alg:rebinding} proposes a fast procedure to update the emission matrix to the new observations by only performing local updates, and to bind the updated emission matrix to the existing schema $T$, defining a new grounded schema $\{T, {\cal C}, E^{\mathrm{rb}}\}$. We call this process \emph{fast rebinding}.

\begin{algorithm}[h!]
\caption{-- Fast rebinding algorithm}
\textbf{Input: } Grounded schema $\{T, {\cal C}, E^0\}$, pseudocount $\epsilon$, prompt $(x_1, \ldots, x_N)$, surprise probability threshold $p_{\text{surprise}}$.\\
\textbf{Output: } Rebound emission matrix $E^{\mathrm{rb}}$
\begin{algorithmic}[1]
    \State{$\tilde{E}^0\propto E^0 + \epsilon$, with normalized rows.}
    \vspace{.5em}
    \State{For timestep $n$, compute $\, p(X_n=j ~|~ x_{\setminus n}) = p({X}_n=j ~|~ x_1, \ldots, x_{n-1}, x_{n+1}, \ldots, x_N), ~ \forall j \le N_\text{obs}$ using $\tilde{E}^0$ for the emission matrix.}
    \vspace{.5em}
    \State{Identify latent states and timesteps that can act as anchors: 
    \newline
    $\mathcal{A} = \left\{(i, n) ~~\big|~~ p(X_n= x_n ~|~ x_{\setminus n}) > p_{\text{surprise}},~  ~\text{and}~ \mathcal{C}(i) = x_n \right\}$}
    \vspace{.5em}
    \State{Identify latent states to be rebound (and their timesteps): 
    \newline
    $\mathcal{R} = \left\{ (i,n) ~~\big|~~ p(X_n=j ~|~ x_{\setminus n}) > p_{\text{surprise}}, ~ j \neq x_n, ~
    (\cdot, n) \notin \mathcal{A} , ~ (i,\cdot) \notin \mathcal{A} ~\text{and}~ \mathcal{C}(i) = j
    \right\}$}
    \vspace{.5em}
    \State{Fix $T$, and use EM to update the emission matrix (initialized with ${E}^0$, and without using any pseudocount) by only using the beliefs for latent states $i$ and timesteps $n$ such that $(i, n) \in \mathcal{R}$.
    }
\end{algorithmic}
\label{alg:rebinding}
\end{algorithm}

Given a prompt $(x_1, \ldots, x_N)$ and a surprise threshold, Algorithm \ref{alg:rebinding} proceeds by identifying the entries of the emission matrix that need to be updated, and then updating those entries using the Expectation-Maximization (EM) algorithm \cite{dempster1977maximum}. The conditional probability $p(X_n=j ~|~ x_{\setminus n})$  of tokens at time step $n$ given all other time steps is used to identify time steps and latent states that are surprising or not.
If the current model predicts an observation with high confidence, and the observation actually occurs, then the latent states corresponding to those observations and time steps should not be updated. These latent states and time steps, called anchors, are identified in Step 3. Latent states corresponding to observations that are incorrectly predicted with high confidence, but are not one of the anchor states, are the candidates for rebinding. These are identified in Step 4.  Finally, Step 5 locally updates the emission matrix using the EM algorithm only on the latent states and time steps identified in step 4. This is a special case of updating the whole emission matrix as described in \cite[][Appendix A.2]{guntupalli2023graph} where the authors re-learn the whole emission matrix of a CSCG by keeping the schema $T$ fixed. In contrast with \cite{guntupalli2023graph}, our EM updates in Step 5, described in detail in Appendix \ref{sec:cscg_learning_local}, only locally update the emission matrix. As a consequence, only a small subset of rows differ between ${E}^0$ and $E^{\mathrm{rb}}$. The rows that are protected correspond either to anchors in the current prompt, or slots not relevant to the current prompt but still possibly relevant to future observations.
The pseudocount used in Step 1 is an uncertainty parameter that lets the model smooth over incorrect observations. More details of this parameter are available in \cite{george2021clone}.

After rebinding, we complete the prompt by performing MAP inference conditioned on the provided prompt in the rebound CSCG. We run the max-product algorithm \cite{pearl1988probabilistic} forward (the backward messages are all uniform) thus generating a series of MAP observations for the tokens following the prompt. We stop once we generate a delimiter token.
See Algorithm \ref{alg:prompt} in Appendix \ref{sec:cscg_appendix} for details.

Section \ref{sec:discussion} discusses how a mechanism similar to Algorithm \ref{alg:rebinding} could be implemented in transformers using buffered inputs and activations.

\section{Outline of the overall argument using CSCG}

\subsection{Context-dependent latent representations and transitive generalization}
\label{sec:retrieval}

\begin{figure}
  \centering
  \includegraphics[width=0.95\textwidth]{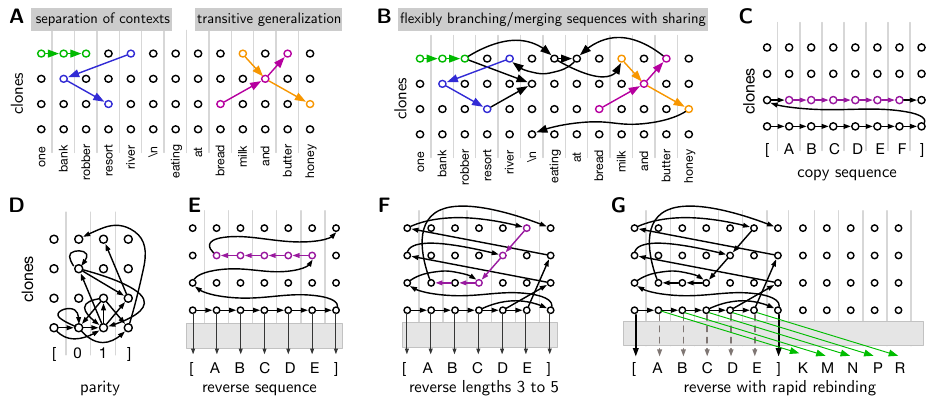}
  \caption{{\bf A}. CSCGs allow both separation of contexts and transitive generalization. The word ``bank'' is wired to different clones that correspond to the different contexts it is used in. If ``milk and honey'', and ``bread and butter'' are seen in training, transitive generalization occurs if they get wired through the same `and' clone, resulting in ``bread and honey'' and ``milk and butter'' being valid sequences. {\bf B}. Probabilistic branching \& merging of sequences. {\bf C} -- {\bf F}. Exemplar CSCG circuits for respectively copying a sequence, parity operation, reversing a list with exactly five elements, reversing lists with a variable number of elements. {\bf G}. Rebinding to new observations: dashed gray arrows correspond to old emissions while green arrows correspond to new rebound emissions.}
  \label{fig:context-examples}
\end{figure}

The clone structure of CSCGs allows context-based separation and appropriate blending for language modeling.
For example, the sense of the word ``bank'' in ``bank robber'' is different from the one in ``river bank''. CSCG learning disambiguates these contexts in the latent space by wiring them to different clones to improve predictive accuracy. In Fig.~\ref{fig:context-examples}A, the sentences ``river bank resort'', and ``one bank robber'' use different clones of ``bank''. Sequences can have probabilistic branching: ``one bank robber'' can terminate at ``$\backslash$n'', or continue to ``eating at river bank resort'' or ``eating bread and honey'', or ``eating bread and butter at river bank resort'' (Fig.~\ref{fig:context-examples}B). CSCGs also allow merging of contexts that result in transitive generalization: even if training data has only the sequences ``bread and butter'', and ``milk and honey'', if they go through the same clone state ``and'', the model will generalize to ``bread and honey'' and ``milk and butter'', assigning non-zero probability to those sequences. Due to the combination of context-sensitive separation and transitivity, related topics, concepts, and algorithms get clustered into sub-networks that pass through the same clones. A prompt's context would activate its sub-network, and transitive generalization allows for prompts that are not exact memorizations. As we show in Section \ref{sec:results}, the Bayesian inference perspective on ICL \cite{xie2021explanation} corresponds to this context-sensitive and transitively generalizing storage and retrieval alone, and is insufficient to explain the ICL properties we consider in the next sections.

\subsection{Learning flexible schemas (template  circuits) and  rebinding}
\label{sec:rebinding}

Just like learning the layout of rooms, CSCG can learn automata circuits \cite{liu2022transformers} for sequence-to-sequence (seq2seq) algorithms. 
See Fig.~\ref{fig:context-examples} for CSCG circuits for parity, copying a sequence, and reversing sequences of multiple lengths. 
The list reversal circuit in Fig.~\ref{fig:context-examples}E is bound to the specific symbols $A, B, C, D, E$ used in training. To be useful as a template, appropriate slots in this graph should be able to bind to arbitrary symbols that occur in context during test time \cite{shanahan2022abstraction,kolodny2015learning}. 
Intuitively, rebinding can be understood as operating based on prediction errors -- if the latent context strongly predicts the latent state corresponding to a time instant, but the real observation at that time is mismatched, rebinding adjusts the emission matrix to wire all the clones of that latent state to the surprising observation. 
This mechanism is formalized in Algorithm \ref{alg:rebinding}.
The rebinding process allows the specific list-reversal circuit in Figure \ref{fig:context-examples}E to become a flexible template with slots that can be dynamically bound to new inputs as required, creating a powerful way to mix and gate previous knowledge with new content.
For example, in the list reversal schema in Fig.~\ref{fig:context-examples}F, tokens ``['', and ``]'' are prior content that detect the beginning and end of the list -- these act as anchors for grounding the schema in the observations. Probabilistic branching based on the end of list token ``]'' allows for length generalization, whereas absorbing arbitrary symbols into the slots corresponding to $A, B, C, D, E$ allows the algorithm to generalize to new symbols. Fig.~\ref{fig:context-examples}G illustrates the outcome of this rebinding mechanism where the slots emitting $A, B, C, D, E$ are respectively rebound to symbols $K, M, N, P, R$ from the input prompt. 
Similarly, in the sentence ``I wrote in a notebook using a dax'', rebinding can absorb the new token ``dax'' into the context by binding it to a clone corresponding to ``pencil'' or ``pen'', and use the new word in those contexts.

\begin{figure}[ht!]
  \centering
  \includegraphics[width=0.95\textwidth]{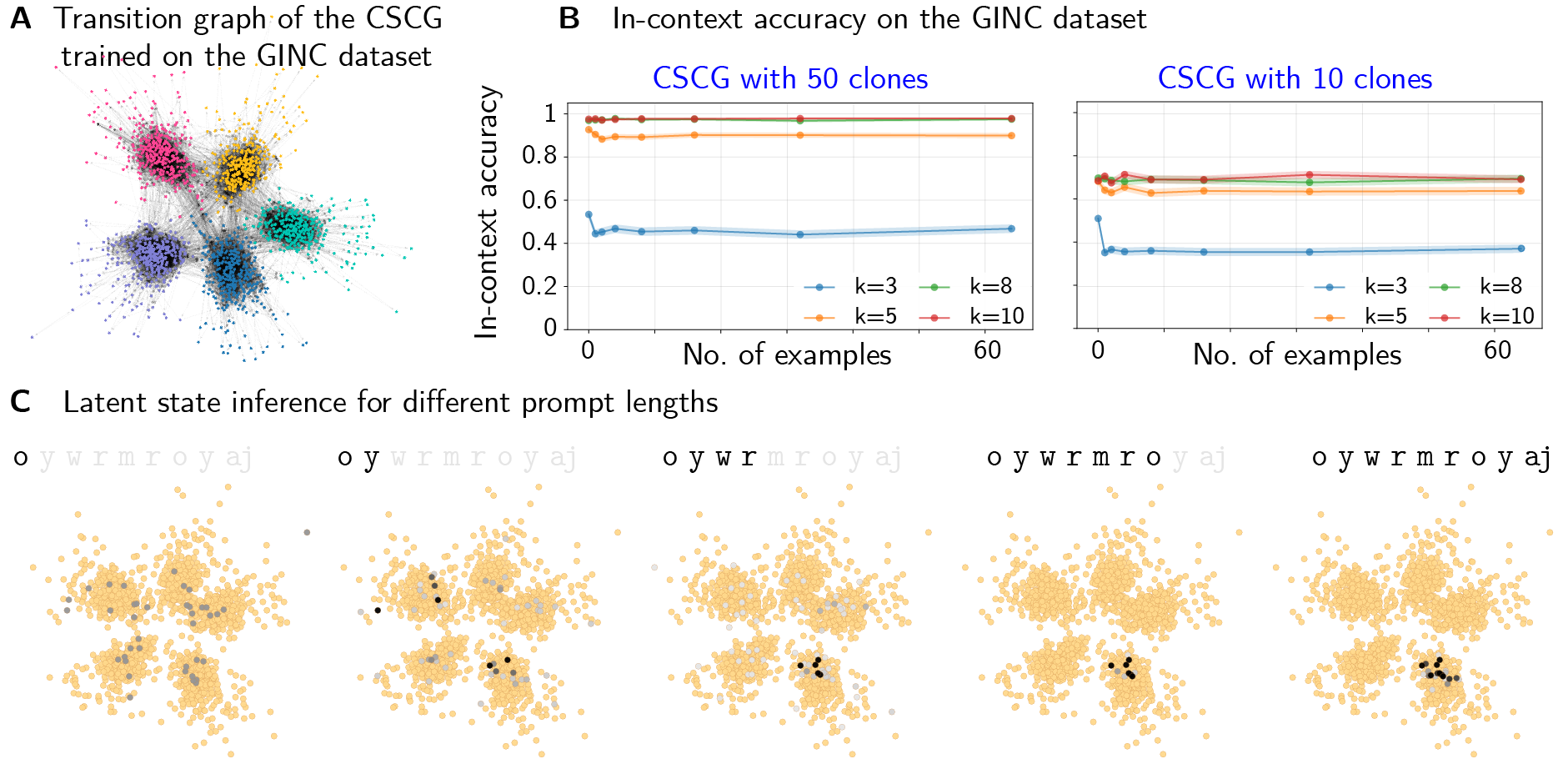}
    \caption{{\bf A}. Visualization of the transition graph of a CSCG with $50$ clones trained on the GINC dataset. The clones are slotted into five clusters, each corresponding to a different ``concept''.  {\bf B}.[Left] In-context accuracy for a CSCG model with $50$ clones averaged over the GINC test dataset from \cite{xie2021explanation}. As in \cite{xie2021explanation} we report the $95\%$ confidence intervals (CIs). For contexts of $8$ and $10$ tokens, the model predicts the most likely next token at least $95\%$ of the time---including in the zero-shot regime. [Right] In-context accuracy decreases when we reduce the number of clones to $10$---for $k\in\{8, 10\}$ it drops from above $95\%$ to below $75\%$. The numerical values are reported in Appendix~\ref{sec:ginc_appendix}, Table~\ref{table:appendix_ginc}. {\bf C}. Decoded latent state distributions (increasing intensities of black for higher density) for a CSCG with 50 clones, for an $n=0$ \& $k=10$ prompt \texttt{o~y~w~r~m~r~o~y~aj}, when truncated to different lengths ($k = {2,3,5,8,10}$). Longer prompts lead to improving latent state estimation---resulting in better concept retrieval, and next token prediction.}
  \label{fig:ginc_results}
\end{figure}

\subsection{Instruction-based or content-based retrieval and completion of tasks}
\label{sec:content-based}

\textbf{Zero-shot task recognition as content-based retrieval using rebinding:} Many striking examples of zero-shot learning involve the recognition of tasks from prompts, and then repeating the recognized task on new inputs. For example, given a prompt ``Input: [p, q, r, s] Output: [p, p, q, q, r, r, s, s]; Input: [l, m, n, o] Output: [l, l, m, m, n, n, o, o]'' LLMs can infer the task as repeating the elements of the sequence, and apply that to complete the output for a new input prompt even when the tokens ``p, q, r, s, l, m, n, o'' were not seen during training, in association with this task. CSCG offers a natural explanation for this using rebinding. Given the prompt, expectation maximization (EM) \cite{dempster1977maximum} simultaneously evaluates the different rebindings to multiple latent algorithm schemas to infer the best binding, which is then applied to complete the query prompt. 
 
\textbf{Instruction-based retrieval: } When algorithms are trained with prefixed language instructions, CSCGs learn instruction sub-networks that directly point to the circuits that represent the algorithms (see Section \ref{sec:seq2seq_algos}). The algorithm can be retrieved by direct prompting with language instructions that can be significantly different from training instructions due to transitive generalization and rebinding. 

\subsection{Emergence}
\label{sec:emergence}
We hypothesize, and empirically demonstrate in Section \ref{sec:results}, that emergence is explainable as the combined effects of the above properties (context-separation, transitive generalization, schema-formation, and rebinding), model capacity, and patterns in the data. Learning the schematic circuits for more complex algorithms or more patterns in the data requires greater model capacity because overparameterization helps in the optimization process. Training on a bigger dataset results in the induction of more templates that might not have occurred in the smaller dataset. 

\section{Results}\label{sec:results}

We substantiate the above argument using empirical results on three datasets: (a) the GINC benchmark introduced in \cite{xie2021explanation}, (b) a suite of algorithm learning tasks that we introduce in our LIALT datasets, and (c) a zero-shot word usage induction task on a CSCG language model. 

\subsection{Context-sensitive retrieval on GINC dataset matches Bayesian inference explanation}\label{sec:ginc_results}

\textbf{Dataset:} 
The GINC dataset, introduced in \cite{xie2021explanation} to study ICL, is generated from a uniform mixture of five factorial HMMs \cite{ghahramani1995factorial}. Each factorial HMM is referred to as a ``concept''. A document is created by concatenating independent sentence samples from a concept. 
The in-context test prompts have a number of examples varying from  $n=0$ to $n=64$: each example can be of lengths $k \in \{3,5,8,10\}$, with $2500$ prompts for each setting $(k,n)$.
Each prompt uniformly selects a concept, samples $n-1$ examples $x^{(1)}_{:k}, \ldots, x^{(n-1)}_{:k}$ of length $k$, and one example $x^{(n)}_{:k-1}$ of length $k-1$. The in-context task is to infer the most likely last token of the last example, i.e., $\argmax_{x_{k-1}^{(n)}} p\left(x_{k-1}^{(n)} | x^{(1)}_{:k}, \ldots, x^{(n-1)}_{:k}, x^{(n)}_{:k-1}\right)$.
Since the vocabulary is shared among different latent concepts, observations in GINC are aliased like in natural language, and solving the task requires the model to disambiguate the aliased observations to correctly infer the latent concepts.  

\textbf{Training:} 
We train a single CSCG with $50$ clones on the GINC dataset for $100$ full-batch EM iterations using a pseudocount \cite{george2021clone} of $\epsilon={10}^{-2}$. Given a test prompt, CSCG infers the most likely hidden sequence for that prompt, then predicts the next most likely observation. 

\textbf{Results:} CSCG learns different latent sub-networks corresponding to the five latent concepts in the GINC dataset ( Fig.~\ref{fig:ginc_results}A), and inference on a supplied prompt retrieves the correct latent sub-network (Fig.~\ref{fig:ginc_results}C). Increasing the prompt length improves the localization of the sub-network and the particular states within the sub-network.   Figure ~\ref{fig:ginc_results}C visualizes the decoded latent state distribution for an example prompt in the zero-shot setting ($n=0$). The decoding starts out uncertain, and improves as the prompt gets longer. This localization (on the graph) results in effective schema retrieval, and hence accurate prompt completion.
Figure ~\ref{fig:ginc_results}B[left] reports the in-context accuracy---defined as the average ratio of correct predictions---for each $(k,n)$ pair of the GINC test set. 
CSCG in-context accuracy matches the patterns exhibited by LSTMs and transformers in \cite{xie2021explanation}, while slightly improving their performance.
Fig.~\ref{fig:ginc_results}B also shows that a CSCG with larger capacity, i.e. with $50$ clones per token, better separates the latent concepts and significantly outperforms a CSCG with only $10$ clones per token.  Fig.~\ref{fig:ginc_results_appendix}[left] in Appendix \ref{sec:ginc_appendix} displays the CSCG in-context confidence: for larger contexts, CSCG is better at disambiguating aliasing and the averaged predictions probabilities are higher. Finally, Fig.~\ref{fig:ginc_results_appendix}[right] shows that similarly to the transformer and LSTM in \cite{xie2021explanation}, CSCG fails at ICL when test prompts are sampled from concepts unseen during training.

The GINC results match the context-based retrieval argument in Section \ref{sec:retrieval}: ICL in this setting is the retrieval of a shared latent concept between the prompt and the model. By using the long-range coherence of concepts in the training documents, the model learns to separate concepts into different latent representations. Despite the train and prompt distribution mismatch \cite{xie2021explanation}, CSCG succeeds at prompt completion because the representation allows transitive mixing.

\subsection{Learning schemas for seq2seq algorithms and generalization using rebinding }
\label{sec:seq2seq_algos}

\begin{figure}
  \centering
  \includegraphics[width=0.95\textwidth]{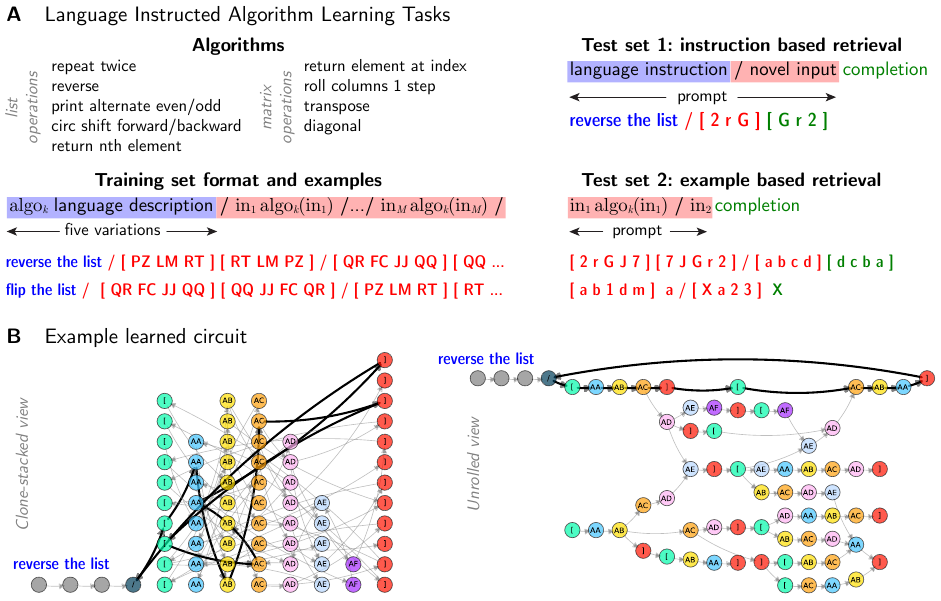}
  \caption{{\bf A}. [Top-left] List and matrix algorithms used in the LIALT dataset. Format of the training set [bottom-left] and examples of the two LIALT test sets [right]. {\bf B}. Example of a learned circuit for the ``reverse'' algorithm, displayed by stacking clones [left] or unrolling them [right].}
  \label{fig:circuits}
\end{figure}

\textbf{Training dataset: }
To test the ability of CSCG to learn \emph{algorithms} that generalize to novel inputs not seen during training, we construct the Language Instructed Algorithm Learning Tasks (LIALT) dataset. The LIALT training set contains demonstrations of $13$ list and matrix algorithms displayed in Fig. \ref{fig:circuits}A[top-left]. A demonstration consists of a multi-word language instruction---each algorithm has five different instructions---followed by $10$ input-output examples of that algorithm. See Tables~\ref{table:appendix_lialt_instructions_list}~\&~\ref{table:appendix_lialt_instructions_matrix} in Appendix \ref{sec:lialt_instructions} for the complete list of instructions used. For each instruction, the dataset contains $20$ demonstrations. Within a demonstration, the language instruction and the examples are separated by a ``/'' delimiter. Demonstrations are separated by a ``$\backslash$n'' delimiter. The input lists and matrices values are created by uniformly sampling from a vocabulary of $676$ tokens, created by random pairings of uppercase letters. List operations examples vary in lengths from $3$ to $6$, and the matrix operations are of sizes $2 \times 2$ or $3 \times 3$. Fig. \ref{fig:circuits}A [bottom-left] shows the training data format.

\begin{figure}
    \centering
    \includegraphics[width=0.95\textwidth]{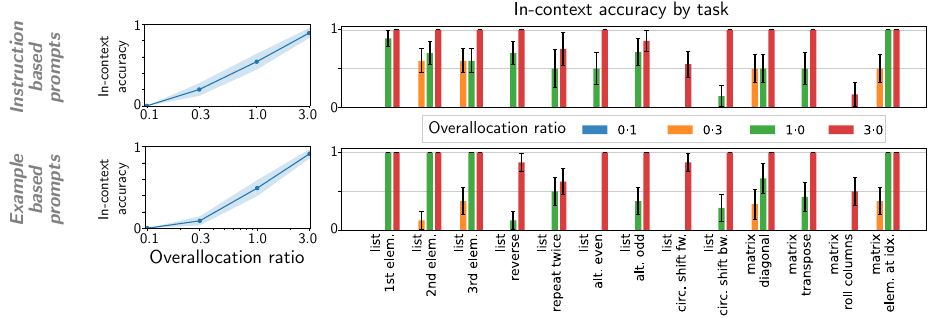}
    \caption{[Left] In-context accuracy (with $95\%$ CIs) after a single EM iteration, as a function of the overallocation ratio for a CSCG trained on LIALT and averaged [top] on the instruction-based LIALT test set [bottom] on the example-based LIALT test set. In-context accuracy increases for CSCGs with larger capacities. [Right] In-context accuracy (with standard errors) per task on the two LIALT test sets: for each task, overparametrization improves performance. Invisible bars indicate zero accuracy for the respective combinations of model and task. All the numerical values are in Appendix \ref{sec:lialt_appendix_results}. Figure~\ref{fig:emergence_em_conv} in the Appendix visualizes the same quantities after EM convergence; the similarity between the two sets of measurements demonstrates that the fast rebinding algorithm is not just localized in its updates, but also rapid.}
    \label{fig:emergence}
    \vspace{-0.5em}
\end{figure}

\begin{figure}[ht!]
    \centering
    \includegraphics[width=0.95\textwidth]{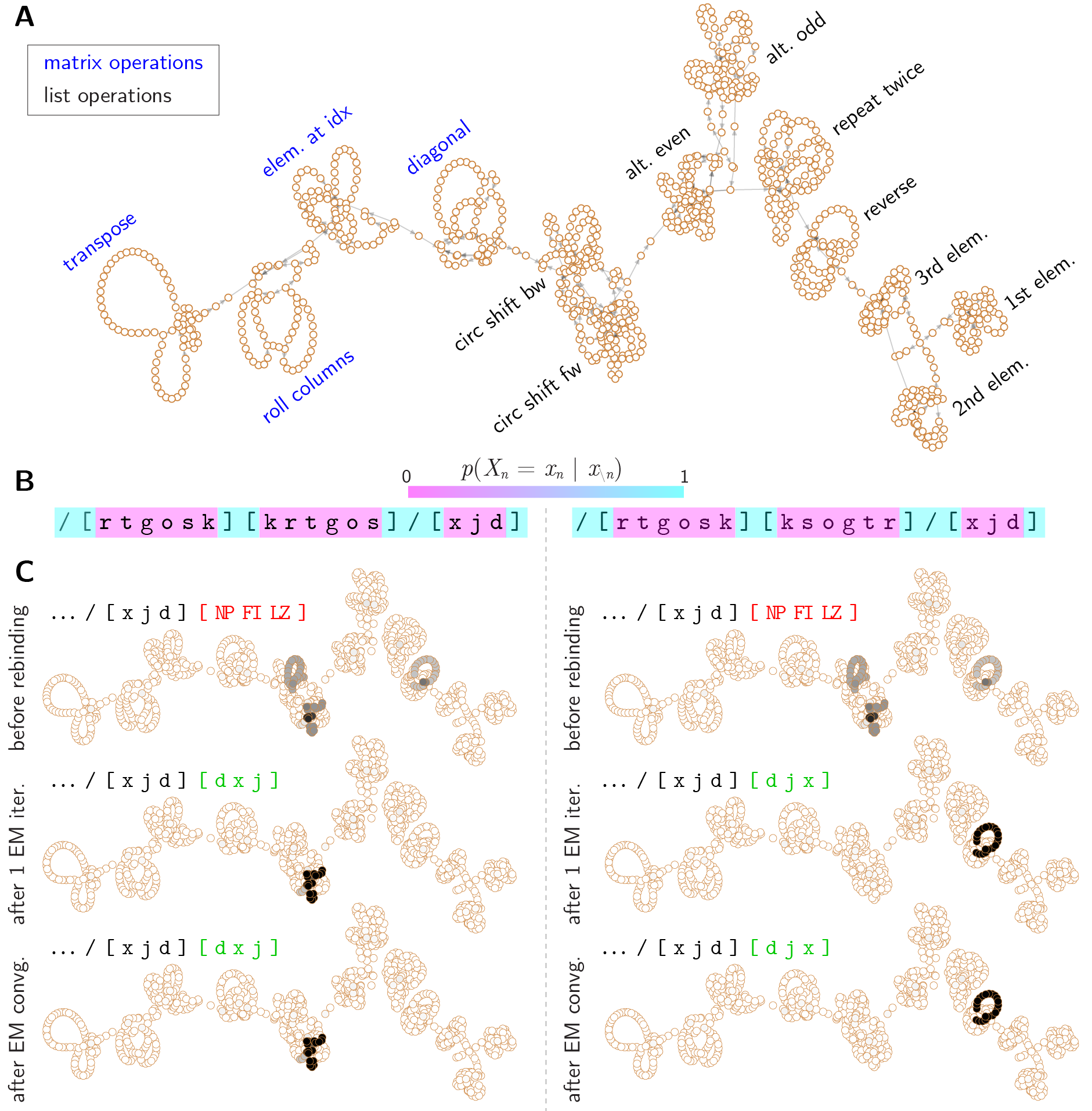}
    \caption{{\bf A.} Transition graph of the CSCG model learned on the LIALT dataset, visualized using the Kamada-Kawai algorithm. {\bf B.} Visualizing the inferred probability of the observation at timestep $n$, conditioned on observations at all other timesteps, before rebinding. This drives the identification of anchors and slots selected for rebinding. {\bf C.} Decoded latent state distributions --- and predicted prompt completions --- for the two different example-based LIALT prompts specified in subfig.~B: (top) before rebinding, (middle) after one iteration of EM, and (bottom) after EM convergence. The left prompt corresponds to the operation of circularly shifting the list forward, and the right prompt corresponds to reversing the list.}
    \label{fig:lialt_kk}
\end{figure}

\textbf{Test dataset: } LIALT has two test datasets, respectively containing: (a) instruction-based retrieval prompts, and (b) example-based retrieval prompts. An instruction-based retrieval test prompt consists of a natural language instruction followed by a single input. An example-based retrieval test prompt consists of a first input-output example of an algorithm, without any natural instruction, followed by a second input. All the lists and matrices in the two test datasets contain novel tokens not seen during training. For both types of prompts, the in-context task is to predict the algorithm's output when applied to the (last) input. Note that for an example-based prompt, CSCG has to infer the algorithm used from the first example. Each test set contains $100$ prompts, constructed by uniformly sampling instructions, and list or matrix tokens. Fig. \ref{fig:circuits}A [right] shows the formats of these two test sets.

\textbf{Training: } 
For each token, a CSCG allocates its number of clones proportionally to the number of distinct contexts in the training data in which it occurs\footnote{As the same token might occur in different contexts in the training data, knowing the context allows predicting the sequence of following tokens, up to the next ``/'' delimiter. }. We parameterize CSCG capacity via this proportionality factor -- the ``overallocation ratio''. 
We train CSCGs for an increasing sequence of overallocation ratios on the training data with $500$ EM iterations and a pseudocount of $\epsilon={10}^{-6}$. After running EM, we run $10$ iterations of Viterbi training \cite{jelinek1976continuous}.

\textbf{Results:} CSCGs with sufficient model capacity successfully learns the algorithms from the training set, and rebinding generalizes those algorithms to novel tokens seen only during test time. Fig.~\ref{fig:circuits}B shows the learned extracted circuit for the list reversal algorithm. Fig.~\ref{fig:emergence}[left] presents the in-context accuracy of CSCGs (using $\epsilon ={10}^{-6}$ and $p_{\text{surprise}} = 0.1$) on the two LIALT test sets: the best performing CSCG (a) successfully rebinds the learned schemas to the test prompts' novel tokens and (b) correctly infers the algorithm from a single input-output pair for example-based prompts. Fig.~\ref{fig:emergence} also shows that model size drives ICL performance [left] even when breaking down the performance by tasks [right].

The learned CSCG (initialized with an overallocation ratio of $3$) is visualized in Fig.~\ref{fig:lialt_stacked} in the Appendix, using stacked clones. Fig.~\ref{fig:lialt_kk}[A] shows the transition graph using the Kamada-Kawai algorithm \cite{kamada1989algorithm}. It reveals thirteen loosely connected clusters corresponding to the thirteen algorithms present in the LIALT dataset. Fig.~\ref{fig:lialt_kk}[B] illustrates the rebinding process, with the decoded distributions over latent states of the learned CSCG model, for two different example-based prompts. Even before any rebinding, the identification of anchors and slots already restricts the decoding to schemas compatible with the prompt \emph{structure}---in this case based on brackets \& delimiters. However, the structure is insufficient to disambiguate completely between the compatible schemas (list operations corresponding to reversal, circular forward shift, and circular backward shift), and both the chosen prompts result in the same latent state distribution. Hence, the decoded distribution after the first E-step localizes to the three compatible schemas. In the M-step that follows, the slots in all three schemas will be rebound for this prompt. At the end of the first EM iteration, the new bindings for slots in the correct schema will be highly certain given the consistent evidence, while inconsistent evidence will lead to uncertain bindings for the other slots. In the E-step of the second iteration, the respective levels of certainty in the bindings then help promote the correct algorithm schema to become the most likely decoding---and complete the prompt appropriately. Note that a single EM step is sufficient to derive the correct rebinding in these examples. Compare Figs.~\ref{fig:emergence}~\&~\ref{fig:emergence_em_conv}, and the tables in Appendix Sec.~\ref{sec:lialt_appendix_results} for how the in-context completion performance after the first EM step in the rebinding process is very similar to that at the end of EM convergence ().

The LIALT results substantiate the arguments we made in Sections \ref{sec:rebinding} and \ref{sec:content-based}. Similar to GINC Section \ref{sec:ginc_results}, the Bayesian inference explanation that infers the latent context based on long-term coherence does not explain the remapping of a latent representation to completely new tokens as required for generalizing the algorithms in LIALT. Without rebinding, a prompt containing a full length example of an algorithm does not localize the correct algorithm schema or produce the correct completion based on inference over the latent states alone (Fig.~\ref{fig:lialt_kk}[B], first row), in contrast to the results from the GINC dataset. The correct algorithm schema cannot be retrieved because the prompt contains novel tokens that were not seen during the training of that algorithm. In contrast, simultaneously inferring the rebinding and the latent states results in accurate retrieval of the algorithm schema and the correct prompt completion (Fig.~\ref{fig:lialt_kk}[B], second and third rows). Utilizing rebinding, CSCG is able to learn seq2seq algorithms and generalize those algorithms to novel tokens that are not encountered in training.

\textbf{Emergence:} ICL performance of CSCG on the LIALT dataset shows characteristics attributed to emergence: accuracy of in-context learning has a clear dependency on the level of overparameterization of CSCG, offering evidence in support of our hypothesis in section \ref{sec:emergence}.

\subsection{Dax test}
\begin{figure}[ht!]
  \centering
  \includegraphics[width=0.95\textwidth]{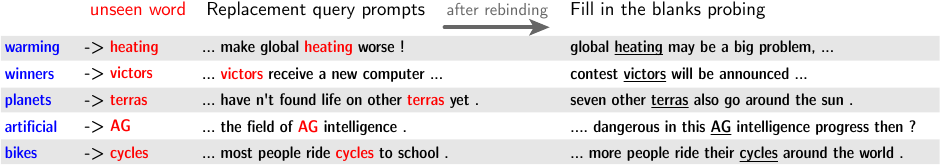}
  \caption{Examples of the dax test performed on a CSCG trained on the PreCo dataset. In each row, the word in red (e.g. ``terras'') in the replacement query prompt is not part of the vocabulary used to train the model. The model absorbs the new token by binding it to the appropriate clones (corresponding to the word in blue, e.g., planets). It can then use the new token in similar contexts as demonstrated in the fill-in-the-blanks probing.}
  \label{fig:dax_test}
\end{figure}

In language, the ``dax'' test \cite{vlach2017remember} is used to demonstrate the capability of a model to absorb the usage of an entirely new word from a single presentation. To test for this capability, we train a CSCG on the PreCo dataset \cite{chen2018preco}, which is a large-scale English dataset for coreference resolution. We then test the model on five word-replaced query prompts, where certain words in the prompts do not appear in the training set. We use Algorithm \ref{alg:rebinding} with $\epsilon ={10}^{-6}$ and $p_{\text{surprise}} = \frac{1}{16}$ to rebind the emission matrix on each of these prompts, each time probing the model for completing a sentence by filling in the blanks (uncertain inputs) using MAP inference. Fig.~\ref{fig:dax_test} shows these results. 

\section{Related work}

\textbf{In-context learning (ICL): } Similar to how humans learn by analogy \cite{winston1980learning} and how synaptic plasticity allows the brain to rapidly adapt to a new task \cite{bittner2017behavioral}, ICL allows a pretrained model to learn a new task given only a few examples. Since \cite{brown2020language} demonstrated the ICL abilities of GPT-3, a substantial body of work has attempted to improve this capability. \cite{wei2022chain, zhou2022least} showed how demonstrations that explicitly guide the reasoning process improve the ICL performance of transformers on new complex tasks. Here, we first clarify some concepts that should not be confused with ICL. We then discuss some works that aim at understanding ICL and the factors that influence it.

\textbf{Supervised learning (SL) and few-shot learning (FSL): } SL approaches learn a mapping that minimizes a loss on the training data: gradient methods are a popular paradigm \cite{bottou2018optimization, ruder2016overview, kingma2014adam}. In the FSL regime, a model learns to rapidly adapt to a new task from a limited number of supervised examples \cite{wang2020generalizing,fei2006one,lake2015human}, and is asked to perform this same task at inference. In contrast, while some works \cite{sanh2021multitask, wei2021finetuned, min2021metaicl, chen2021meta} finetuned the pretrained model for ICL, the new ICL tasks are only revealed during inference. \cite{wei2021finetuned, sanh2021multitask} showed that finetuning transformers on instructions improves their ICL performance.

\textbf{Meta-learning: } The meta-learning paradigm aims at learning to adapt to a new task with only a few examples \cite{naik1992meta, ravi2017optimization, hochreiter2001learning} by using multiple learning experiences. In contrast, ICL directly emerges from the pretrained model. \cite{min2021metaicl, chen2021meta} proposed a meta-learning framework for ICL where the model is fine-tuned: it learns to leverage few-shot examples and to adapt to new tasks at inference time.

\textbf{How ICL works: } Several studies have highlighted the role of the training data in ICL. \cite{chan2022data} showed that ICL emerges when the training data has (a) examples appearing in clusters, and (b) a large number of rare classes. \cite{xie2021explanation} explained ICL as implicit Bayesian inference and constructed the GINC dataset (see Section \ref{sec:ginc_results}) for demonstrating ICL. Some works have also attempted to understand the mechanics of ICL. \cite{litransformers} abstracted ICL as an algorithm learning problem and found that a transformer can implicitly infer a hypothesis
function. Similarly, \cite{garg2022can} showed that a transformer can be trained to perform ICL of unseen linear functions, with performance comparable to the optimal least squares estimator. \cite{akyurek2022learning} showed that, in the linear case, transformers (a) implicitly implement gradient descent and (b) train an implicit linear model on the ICL examples. \cite{dai2022can} proposed a dual between transformer attention and gradient methods and suggested pretrained models are meta-optimizer. They presented ICL as implicit finetuning, where the forward pass on the demonstrative examples produces meta-gradients. Finally, \cite{olsson2022context} showed the existence of ``induction heads'' in transformers, that emerge during training, copy previous patterns, and drive ICL capacities.

\textbf{What influences ICL: }
\cite{li2022systematic} proposed a substitute for the positional encoding, and demonstrated how transformers can learn schemas for algorithmic tasks and generalize to test sequences longer than any seen during training. \cite{shin2022effect} highlighted that (a) ICL can emerge when a model is trained on a combination of multiple corpora (b) low perplexity and ICL performance do not always correlate. \cite{brown2020language, xie2021explanation} found that transformers' ICL performance improves with (a) the model size and (b) the number of demonstrative examples. Similarly, \cite{wei2022emergent} indicated that ICL ``emerges'' when model size increases. Some works have studied the influence of the demonstration samples in ICL. \cite{zhao2021calibrate, lu2021fantastically} found that ICL is highly unstable and is influenced by the prompting template, the selection of in-context examples, and the order of the examples. \cite{min2022rethinking} showed that the ICL performance is driven by the exposure to (a) the label space, (b) the input distribution, and (c) the overall format of the sequence. Similarly, \cite{liu2021makes} found that selecting ICL examples with closer embeddings to ICL test sample improves ICL performance, and \cite{lampinen2022can} found that adding explanations in-context improves performance. Finally, \cite{schaeffer2023emergent} recently claimed that the sharp emergence of ICL in larger models might be an artifact of the metrics, not a fundamental property of the model.

\section{Discussion}
\label{sec:discussion}

\begin{figure}[ht!]
  \centering
  \includegraphics[width=0.95\textwidth]{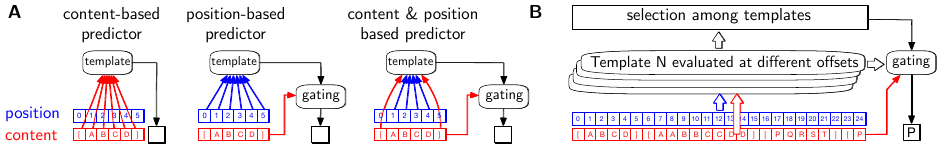}
  \caption{{\bf A}. Learned templates in a transformer could involve content, position, or a mix of both. {\bf B}. Activations in the forward pass of a transformer could be selected among pre-learned templates that mix content and position to achieve ICL without weight changes.}
  \label{fig:gating}
\end{figure}

Several prior explanations for ICL, including Bayesian inference \cite{xie2021explanation} and implicit gradient descent \cite{dai2022can},  have treated all content as the same without separating out the contributions of contents vis-a-vis positions. As we showed, generalizing many algorithms requires templates that can gate content based on patterns in both content and positions. We argue that theories might need to emphasize this distinction (see Fig. \ref{fig:gating}A) to fully understand the inductive biases behind ICL.

Rebinding in CSCG requires localized weight changes to the emission matrix. Unlike CSCGs, transformers buffer the inputs and use attention for the purpose of gating. Thanks to the positional encoding, transformers are able to gate both by location and content. Since attention is implemented as part of the forward pass, the slotting process lives in the space of activations (as opposed to requiring weight modification). This is only a superficial difference: it is possible to temporally unroll the rebinding process in CSCGs, to provide an algorithm for the same output, but with fixed weights. Our conjecture is that layers of the transformer implement multiple mixed templates of positions and content, that are evaluated at different offsets of a prompt. The template assembly that can autoregressively match the prompt wins out the competition to gate the content. 

Although we have illustrated here the concept of rebinding to attach new symbols to existing slots, rebinding can also be done ``through time’’. Just like in our examples the connections among the clones ($T$) remained unchanged, but the connection between the clones and the symbols ($E$) was rebound based on the prompt, it is possible to do the converse, rebind the connections between the clones ($T$) while keeping the connection to the symbols ($E$) unchanged. For instance, there could be a schema within $T$ that recognizes an instruction, and another schema that executes it, triggered by the last clone of the instruction recognizer. If we want a known instruction and all its variants to execute a different task, it is possible to simply rebind the last clone of the instruction recognizer to the trigger of a new task. Both types of rebinding can be combined. Transformers also show this type of generalization, but we leave the details of this approach for future work.

\section*{Acknowledgements}
We thank Stephanie Chan, Andrew Lampinen, Anirudh Goyal, Dharshan Kumaran, Neel Nanda and Guangyao Zhou for helpful discussions and comments on the draft.

\printbibliography

\newpage
\appendix
\section{Locally updating the emission matrix with the transition matrix fixed}\label{sec:cscg_learning_local}

We reuse the same notations as in \cite{guntupalli2023graph}, Appendix A.2. The authors describe the EM agorithm for learning the emission matrix of a CSCG with a fixed transition matrix. In particular their M step defines the new emission matrix as:
\begin{equation*}
E(j) = \sum_{n=1}^{N} 1_{X_{n} = j} ~\gamma(n) \oslash \sum_{n=1}^{N} \gamma(n), ~ \forall j
\end{equation*}
where $E(j)$ is a column of the emission matrix corresponding to the emission $j$, $1_{X_{n} = j}$ is an indicator function, $\oslash$ is the element-wise division and $\gamma(n)$ is derived by the authors from the forward and backward probabilities. The $i$ entry of the vector $E(j)$ is then defined as:
$$
E_{ij} 
= \frac{\sum_{n=1}^{N} 1_{X_{n} = j} ~\gamma_i(n)}{\sum_{n=1}^{N} \gamma_i(n)}.
$$
In contrast, in Section \ref{sec:alg_rebinding}, Step 5 of Algorithm 1 only updates the row $i$ for which we can find a pair $(i,n) \in \mathcal{R}$, by only using the beliefs at timestep $n$. For this row, the $j$th entry becomes:
$$
E_{ij} 
= \frac{\sum_{n: ~ (i,n) \in \mathcal{R}} 1_{X_{n} = j}~ \gamma_i(n)}{\sum_{n: ~ (i,n) \in \mathcal{R}} \gamma_i(n)}.
$$

\section{Prompt completion algorithm}\label{sec:cscg_appendix}

Algorithm \ref{alg:prompt} describes the prompt completion algorithm introduced in Section \ref{sec:alg_rebinding}. It implicitly considers a single action, which takes the next sequence element. 
\begin{algorithm}[htbp]
\caption{-- Prompt completion}
\textbf{Input: } Grounded schema $\{T, {\cal C}, E^{\mathrm{rb}}\}$ with rebound CSCG emission matrix $E^{\mathrm{rb}}$, delimiter token $x_{\emptyset}$, prompt $x^{(\textrm{prompt})} = (x_1,\ldots,x_N)$\\
\textbf{Output: } A completed prompt $x^{(\text{completed})}=(x_1,\ldots,x_N, x_{N+1}, \ldots, x_{N+P}=x_{\emptyset})$
\begin{algorithmic}[1]
    \State{Run max-product for MAP inference and return $z^{\text{MAP}} =(z_1, \ldots, z_N) = \argmax_z p(z|x^{(\text{prompt})})$.}
    \State{Set $\ell=0$. While $x_{N + \ell}\ne x_{\emptyset}$, increment $\ell \leftarrow \ell + 1$ and sample the next most likely observation:
    \newline
    $~~~~~z_{N + \ell} \in \argmax_j T_{z_{N+ \ell - 1}, ~ j}$ and $x_{N + \ell} \in \argmax_j E^{\mathrm{rb}}_{z_{N+ \ell}, ~ j}$.}
\end{algorithmic}
\label{alg:prompt}
\end{algorithm}

\section{Additional materials for the GINC dataset}\label{sec:ginc_appendix}
First, we present two additional plots for the GINC experiment.
\begin{figure}[h!]
  \centering
  \includegraphics[width=.88 \textwidth]{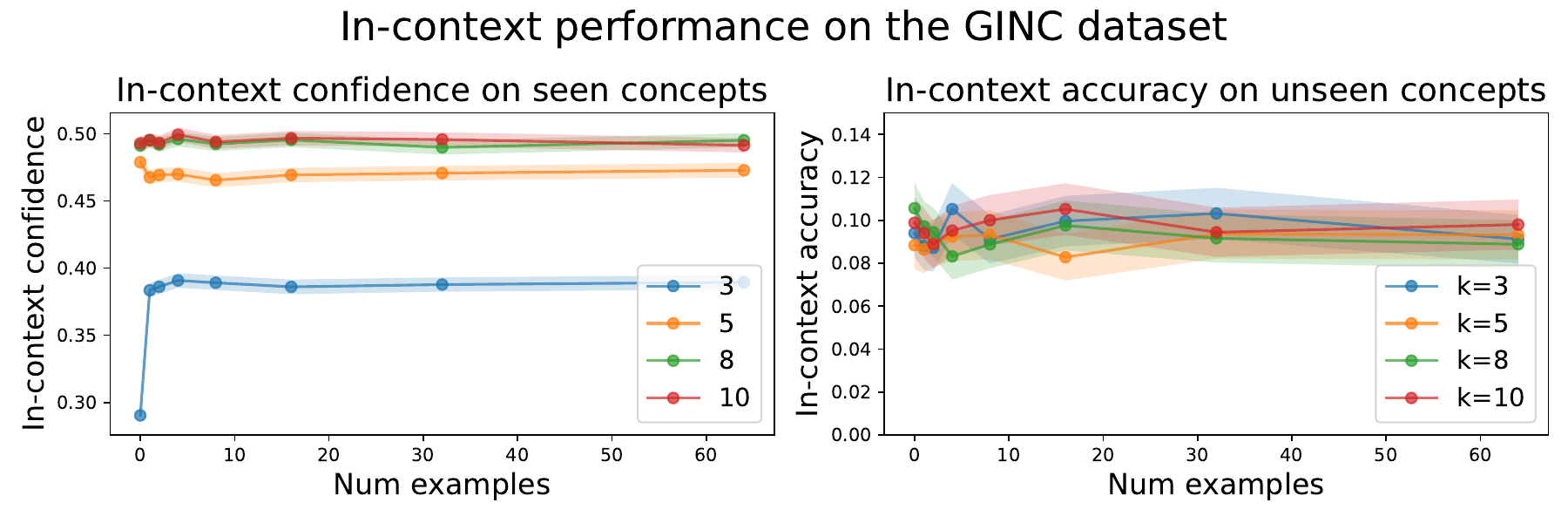}
  \caption{[Left] In-context confidence for the CSCG with $50$ clones on the GINC test dataset, defined as the averaged probability of the predictions. For larger values of $k$, CSCG correctly infers the context of the aliased observations and is more confident in its predictions. [Right] Similar to the transformer and LSTM reported in \cite{xie2021explanation}, CSCG fails to extrapolate and has a low in-context accuracy when the test prompts are sampled from five novel concepts, unseen during training.}
  \label{fig:ginc_results_appendix}
\end{figure}

Second, we present the table of results associated with Fig. \ref{fig:ginc_results} for the CSCGs with $10$ and $50$ clones.
\begin{table*}[!ht]
\footnotesize
\centering
\resizebox{.8 \textwidth}{!}{
\begin{tabular}{P{0.15\textwidth}P{0.2\textwidth}P{0.25\textwidth}P{0.25\textwidth}}
\toprule
Context length & No. of examples & CSCG with $10$ clones & CSCG with  $50$ clones \\
\toprule
\toprule
$3$ & $0$ &  $0.509~(0.020)$ & $0.534~(0.020)$\\
& $1$ &  $0.351~(0.019)$ & $0.445~(0.019)$\\
& $2$ &  $0.366~(0.019)$ & $0.453~(0.020)$\\
& $4$ &  $0.356~(0.019)$ & $0.468~(0.020)$\\
& $8$ &  $0.360~(0.019)$ & $0.454~(0.020)$\\
& $16$ &  $0.354~(0.019)$ & $0.460~(0.020)$\\
& $32$ &  $0.354~(0.019)$ & $0.441~(0.0219)$\\
& $64$ &  $0.369~(0.019)$ & $0.468~(0.020)$\\
\midrule
$5$ & $0$ &  $0.682~(0.018)$ & $0.927~(0.010)$\\
& $1$ &  $0.640~(0.019)$ & $0.927~(0.012)$\\
& $2$ &  $0.629~(0.019)$ & $0.904~(0.012)$\\
& $4$ &  $0.654~(0.019)$ & $0.883~(0.013)$\\
& $8$ &  $0.627~(0.019)$ & $0.894~(0.012)$\\
& $16$ &  $0.637~(0.019)$ & $0.902~(0.012)$\\
& $32$ &  $0.634~(0.019)$ & $0.901~(0.012)$\\
& $64$ &  $0.637~(0.019)$ & $0.899~(0.012)$\\
\midrule
$8$ & $0$ &  $0.696~(0.018)$ & $0.969~(0.007)$\\
& $1$ &  $0.694~(0.018)$ & $0.972~(0.007)$\\
& $2$ &  $0.686~(0.018)$ & $0.972~(0.006)$\\
& $4$ &  $0.681~(0.018)$ & $0.978~(0.006)$\\
& $8$ &  $0.690~(0.018)$ & $0.973~(0.006)$\\
& $16$ &  $0.686~(0.018)$ & $0.975~(0.006)$\\
& $32$ &  $0.676~(0.018)$ & $0.968~(0.006)$\\
& $64$ &  $0.694~(0.018)$ & $0.975~(0.007)$\\
\midrule
$10$ & $0$ &  $0.684~(0.018)$ & $0.975~(0.006)$\\
& $1$ &  $0.705~(0.018)$ & $0.977~(0.006)$\\
& $2$ &  $0.674~(0.018)$ & $0.971~(0.006)$\\
& $4$ &  $0.713~(0.018)$ & $0.974~(0.006)$\\
& $8$ &  $0.690~(0.018)$ & $0.977~(0.006)$\\
& $16$ &  $0.689~(0.018)$ & $0.977~(0.006)$\\
& $32$ &  $0.712~(0.018)$ & $0.978~(0.006)$\\
& $64$ &  $0.690~(0.018)$ & $0.978~(0.006)$\\
\bottomrule
\end{tabular}}
\caption{In-context accuracy for a CSCG with $10$ clones and a CSCG $50$ clones trained on the GINC dataset, averaged (with $95\%$ confidence intervals) on each each pair $(k, n)$ of context length and number of examples $n$ of the GINC test set.}
\label{table:appendix_ginc}
\end{table*}

\textbf{CSCG performs better on zero-shot prompts than on few-shot prompts: }
We observe that, for short contexts, CSCG in-context accuracy is higher on zero-shot prompts $n=0$ than on few-shot prompts $n=1, 2, \ldots$. We hypothesize that the difference between the training and the prompt distributions creates a gap that lowers few-shot in-context accuracy. The performance gap disappears for larger contexts $k\in\{8,10\}$ as they ``overpower'' the train-test distribution divergence. \cite{xie2021explanation} made a similar observation for transformers. However, their performance gap was also observable for larger contexts.

\section{Additional materials for the LIALT dataset}\label{sec:lialt_appendix}

\subsection{Natural language instructions}\label{sec:lialt_instructions}

Tables \ref{table:appendix_lialt_instructions_list} and \ref{table:appendix_lialt_instructions_matrix} present the natural language instructions respectively used for the nine list algorithms and four matrix algorithms of the LIALT dataset. Language instructions are grouped in clusters of five: all five instructions within one cluster describe to the same algorithm. As described in the main text, each demonstration of the LIALT training and first test set uniformly selects one instruction.
\begin{table*}[!ht]
\footnotesize
\centering
\resizebox{\textwidth}{!}{%
\begin{tabular}{p{.65\textwidth} p{.01\textwidth} p{.65\textwidth}}
\cmidrule(lr){1-1} \cmidrule(lr){3-3}
\texttt{``find the element at index zero of the list''} &&\texttt{``print the element at index one of the list''}\\
\texttt{``print the first element from the list''} &&\texttt{``find the second element from the list''}\\
\texttt{``return the leading element from the list''} &&\texttt{``retrieve the second element from the list''}\\
\texttt{``find the head element from the list''} &&\texttt{``locate the second item from the list''}\\
\texttt{`retrieve the starting element from the list''} &&\texttt{``return the element in second place from the list''}\\
\cmidrule(lr){1-1} \cmidrule(lr){3-3}
\texttt{``print the element at index two of the list''} &&\texttt{``reverse the list''}\\
\texttt{``find the third element from the list''} &&\texttt{``mirror the list''}\\
\texttt{``locate the third element from the list''} &&\texttt{``flip the list''}\\
\texttt{``output the third item from the list''} &&\texttt{``flip the order of the list''}\\
\texttt{``return the element in third place from the list''} &&\texttt{``reverse the order of the items in the list''}\\
\cmidrule(lr){1-1} \cmidrule(lr){3-3}
\texttt{``duplicate each list item''} &&\texttt{``rotate the list elements one place forward''}\\
\texttt{``replicate every element in the list''} &&\texttt{``roll the list elements one position to the right''}\\
\texttt{``make a copy of each element in the list''} &&\texttt{``switch the items of the list one position forward''}\\
\texttt{``clone each element in the list''} &&\texttt{``advance the list elements one index forward''}\\
\texttt{``create a second instance of every element in the list''} &&\texttt{``move the list elements one position forward''}\\
\cmidrule(lr){1-1} \cmidrule(lr){3-3}
\texttt{``print every other member in the list starting with the second member''}&&\texttt{``print every other member in the list starting with the first member''}\\
\texttt{``retrieve alternate items in the list starting with the second item''} &&\texttt{``find alternate elements in the list beginning with the first element''}\\
\texttt{``return every other object in the list starting with the second object''}  &&\texttt{``print every second item in the list, starting with the first element''}\\
\texttt{``retrieve every other entry in the list starting with the second entry''} &&\texttt{``output every second element in the list, starting from the first element''}\\
\texttt{``output odd indexed elements''} &&\texttt{``output even indexed elements''}\\
\cmidrule(lr){1-1} \cmidrule(lr){3-3}
\texttt{``rotate the list elements one place backward''}\\
\texttt{``move the list elements one position to the left''}\\
\texttt{``change the items of the list one position backward''}\\
\texttt{``displace the elements of the list one index backward''}\\
\texttt{``roll the list items one position backward''}\\
\cmidrule(lr){1-1} 
\end{tabular}
}
\caption{Natural language instructions for the list algorithms used in the LIALT dataset}
\label{table:appendix_lialt_instructions_list}
\end{table*}
\begin{table*}[h!]
\footnotesize
\centering
\resizebox{\textwidth}{!}{%
\begin{tabular}{p{.65\textwidth} p{.01\textwidth} p{.65\textwidth}}
\cmidrule(lr){1-1} \cmidrule(lr){3-3}
\texttt{``return the matrix diagonal''} &&\texttt{``return the matrix transpose''}\\
\texttt{``collect the diagonal values of the matrix''} &&\texttt{``retrieve the transpose of the matrix''}\\
\texttt{``retrieve the diagonal elements of the matrix''} &&\texttt{``get the transposed matrix''}\\
\texttt{``return the diagonal entries of the matrix''} &&\texttt{``compute the transposed form of the matrix''}\\
\texttt{``fetch the diagonal items of the matrix''} &&\texttt{``derive the transpose matrix''}\\
\cmidrule(lr){1-1} \cmidrule(lr){3-3}
\texttt{``roll the columns of the matrix to the right''} &&\texttt{``find the matrix element in the second row and second column''}\\
\texttt{``rotate the matrix columns to the right''} &&\texttt{`find the value in the second row and second column of the matrix''}\\
\texttt{``move the matrix columns to the right''} &&\texttt{``fetch the matrix element located in row 2 and column 2''}\\
\texttt{``shift the columns of the matrix to the right''} &&\texttt{``print the value at 2 2 in the matrix''}\\
\texttt{``spin the matrix columns to the right''} &&\texttt{``retrieve the matrix element at 2 2''}\\
\cmidrule(lr){1-1} \cmidrule(lr){3-3}
\end{tabular}}
\caption{Natural language instructions for the matrix algorithms used in the LIALT dataset}
\label{table:appendix_lialt_instructions_matrix}
\end{table*}

\subsection{Learned CSCG model}

Our next Figure \ref{fig:lialt_stacked} displays the transition graph of the CSCG model trained on the LIALT dataset with an overallocation ratio of 3, with stacked clones for each symbol.
\begin{figure}[htb!]
  \centering
  \includegraphics[height=.9\textheight]{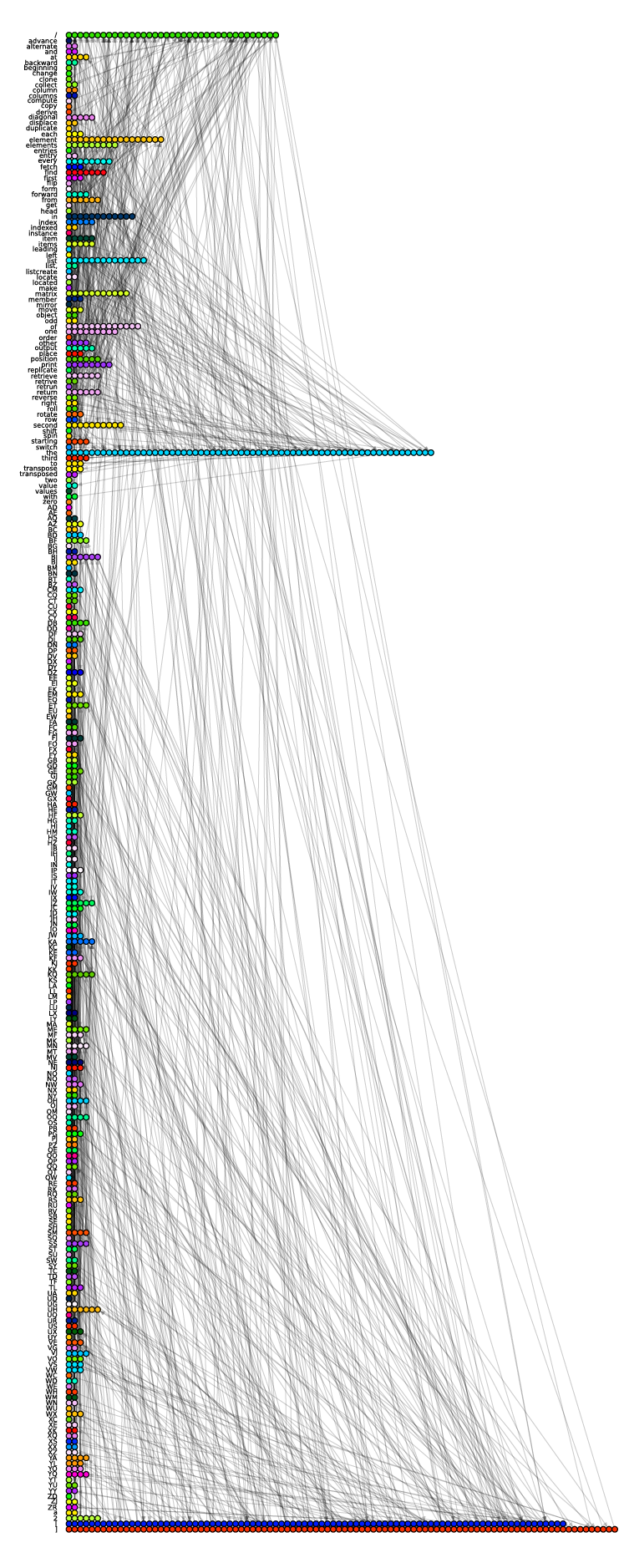}
  \caption{CSCG model learned on the LIALT dataset, visualized with stacked clones.}
  \label{fig:lialt_stacked}
\end{figure}

\newpage
\subsection{Results on the LIALT dataset}
\label{sec:lialt_appendix_results}

\subsubsection{After a single EM iteration}
Presented below are the tables of results associated with Fig.~\ref{fig:emergence}. Table \ref{table:appendix_lialt_emergence} contains the in-context accuracies averaged on the entire test set, Table \ref{table:appendix_lialt_bytask_instructions} contains the in-context accuracies per task on instructions-based prompts, and Table \ref{table:appendix_lialt_bytask_examples} contains the in-context accuracies per task on example-based prompts.
\begin{table*}[htb!]
\footnotesize
\centering
\resizebox{.8 \textwidth}{!}{%
\begin{tabular}{P{0.3\textwidth}P{0.3\textwidth}P{0.3\textwidth}}
\toprule
 Overallocation ratio & Instruction-based prompts & Example-based prompts \\
\midrule
0.1 &               0.00 (0.00) &           0.00 (0.00) \\
0.3 &               0.20 (0.08) &           0.09 (0.06) \\
1.0 &               0.54 (0.10) &           0.49 (0.10) \\
3.0 &               0.89 (0.06) &           0.91 (0.06) \\
\bottomrule
\end{tabular}}
\caption{Average in-context accuracy of each CSCG model---with $95\%$ confidence intervals---as a function of CSCG overallocation on both (a) the instruction-based LILAT test set and (b) the example-based LIALT test set.}
\label{table:appendix_lialt_emergence}
\end{table*}
\begin{table*}[htb!]
\footnotesize
\centering
\resizebox{.8 \textwidth}{!}{
\begin{tabular}{p{0.3\textwidth}P{0.15\textwidth}P{0.15\textwidth}P{0.15\textwidth}P{0.15\textwidth}}
\toprule
& \multicolumn{4}{c}{Overallocation ratio} \\
 \cmidrule(lr){2-5}
Task &          0.1 &          0.3 &          1.0 &          3.0 \\
\midrule
       list 1st elem. &  0.00 (0.00) &  0.00 (0.00) &  0.89 (0.10) &  1.00 (0.00) \\
       list 2nd elem. &  0.00 (0.00) &  0.60 (0.15) &  0.70 (0.14) &  1.00 (0.00) \\
       list 3rd elem. &  0.00 (0.00) &  0.60 (0.15) &  0.60 (0.15) &  1.00 (0.00) \\
         list reverse &  0.00 (0.00) &  0.00 (0.00) &  0.70 (0.14) &  1.00 (0.00) \\
    list repeat twice &  0.00 (0.00) &  0.00 (0.00) &  0.50 (0.25) &  0.75 (0.22) \\
       list alt. even &  0.00 (0.00) &  0.00 (0.00) &  0.50 (0.20) &  1.00 (0.00) \\
        list alt. odd &  0.00 (0.00) &  0.00 (0.00) &  0.71 (0.17) &  0.86 (0.13) \\
 list circ. shift fw. &  0.00 (0.00) &  0.00 (0.00) &  0.00 (0.00) &  0.56 (0.17) \\
 list circ. shift bw. &  0.00 (0.00) &  0.00 (0.00) &  0.14 (0.13) &  1.00 (0.00) \\
      matrix diagonal &  0.00 (0.00) &  0.50 (0.18) &  0.50 (0.18) &  1.00 (0.00) \\
     matrix transpose &  0.00 (0.00) &  0.00 (0.00) &  0.50 (0.20) &  1.00 (0.00) \\
  matrix roll columns &  0.00 (0.00) &  0.00 (0.00) &  0.00 (0.00) &  0.17 (0.15) \\
 matrix elem. at idx. &  0.00 (0.00) &  0.50 (0.18) &  1.00 (0.00) &  1.00 (0.00) \\
\bottomrule
\end{tabular}}
\caption{Average in-context accuracy by task---with standard errors---as a function of CSCG overallocation on instruction-based prompts.}
\label{table:appendix_lialt_bytask_instructions}
\end{table*}
\begin{table*}[htb!]
\footnotesize
\centering
\resizebox{.8 \textwidth}{!}{
\begin{tabular}{p{0.3\textwidth}P{0.15\textwidth}P{0.15\textwidth}P{0.15\textwidth}P{0.15\textwidth}}
\toprule
& \multicolumn{4}{c}{Overallocation ratio} \\
 \cmidrule(lr){2-5}
Task &          $0.1$ &          $0.3$ &          $1.0$ &          $3.0$ \\
\midrule
       list 1st elem. &  0.00 (0.00) &  0.00 (0.00) &  1.00 (0.00) &  1.00 (0.00) \\
       list 2nd elem. &  0.00 (0.00) &  0.12 (0.12) &  1.00 (0.00) &  1.00 (0.00) \\
       list 3rd elem. &  0.00 (0.00) &  0.38 (0.17) &  1.00 (0.00) &  1.00 (0.00) \\
         list reverse &  0.00 (0.00) &  0.00 (0.00) &  0.12 (0.12) &  0.88 (0.12) \\
    list repeat twice &  0.00 (0.00) &  0.00 (0.00) &  0.50 (0.18) &  0.62 (0.17) \\
       list alt. even &  0.00 (0.00) &  0.00 (0.00) &  0.00 (0.00) &  1.00 (0.00) \\
        list alt. odd &  0.00 (0.00) &  0.00 (0.00) &  0.38 (0.17) &  1.00 (0.00) \\
 list circ. shift fw. &  0.00 (0.00) &  0.00 (0.00) &  0.00 (0.00) &  0.88 (0.12) \\
 list circ. shift bw. &  0.00 (0.00) &  0.00 (0.00) &  0.29 (0.17) &  1.00 (0.00) \\
      matrix diagonal &  0.00 (0.00) &  0.33 (0.19) &  0.67 (0.19) &  1.00 (0.00) \\
     matrix transpose &  0.00 (0.00) &  0.00 (0.00) &  0.43 (0.19) &  1.00 (0.00) \\
  matrix roll columns &  0.00 (0.00) &  0.00 (0.00) &  0.00 (0.00) &  0.50 (0.18) \\
 matrix elem. at idx. &  0.00 (0.00) &  0.38 (0.17) &  1.00 (0.00) &  1.00 (0.00) \\
\bottomrule
\end{tabular}}
\caption{Average in-context accuracy by task---with standard errors---as a function of CSCG overallocation on example-based prompts.}
\label{table:appendix_lialt_bytask_examples}
\end{table*}

\subsubsection{After EM convergence}

\begin{figure}[h!]
    \centering
    \includegraphics[width=0.95\textwidth]{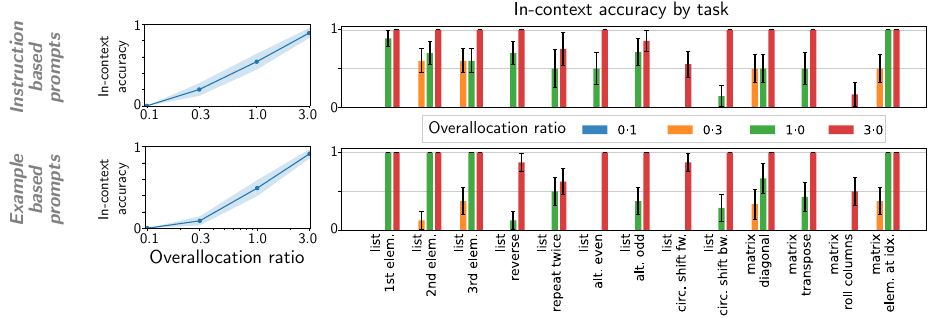}
    \caption{[Left] In-context accuracy (with $95\%$ CIs) after EM convergence, as a function of the overallocation ratio for a CSCG trained on LIALT and averaged [top] on the instruction-based LIALT test set [bottom] on the example-based LIALT test set. In-context accuracy increases for CSCGs with larger capacities. [Right] In-context accuracy (with standard errors) per task on the two LIALT test sets: for each task, overparametrization improves performance. All the numerical values are in Appendix \ref{sec:lialt_appendix_results}. Invisible bars indicate zero accuracy for the respective combination of model and task.}
    \label{fig:emergence_em_conv}
    \vspace{-0.5em}
\end{figure}

Fig.~\ref{fig:emergence_em_conv} presents the analogue of Fig.~\ref{fig:emergence} but after the EM algorithm in the rebinding process has converged. We note that the results are mostly identical. Table \ref{table:appendix_lialt_emergence_em_conv} contains the in-context accuracies averaged on the entire test set, Table \ref{table:appendix_lialt_bytask_instructions_em_conv} contains the in-context accuracies per task on instructions-based prompts, and Table \ref{table:appendix_lialt_bytask_examples_em_conv} contains the in-context accuracies per task on example-based prompts.

\begin{table*}[h!]
\footnotesize
\centering
\resizebox{.8 \textwidth}{!}{%
\begin{tabular}{p{0.3\textwidth}p{0.3\textwidth}p{0.3\textwidth}}
\toprule
 Overallocation ratio & Instruction-based prompts & Example-based prompts \\
\midrule
0.1 &               0.00 (0.00) &           0.00 (0.00) \\
0.3 &               0.16 (0.07) &           0.11 (0.06) \\
1.0 &               0.54 (0.10) &           0.49 (0.10) \\
3.0 &               0.89 (0.06) &           0.93 (0.05) \\
\bottomrule
\end{tabular}}
\caption{Average in-context accuracy of each CSCG model---with $95\%$ confidence intervals---as a function of CSCG overallocation on both (a) the instruction-based LILAT test set and (b) the example-based LIALT test set.}
\label{table:appendix_lialt_emergence_em_conv}
\end{table*}
\begin{table*}[h!]
\footnotesize
\centering
\resizebox{.8 \textwidth}{!}{
\begin{tabular}{p{0.3\textwidth}P{0.15\textwidth}P{0.15\textwidth}P{0.15\textwidth}P{0.15\textwidth}}
\toprule
& \multicolumn{4}{c}{Overallocation ratio} \\
 \cmidrule(lr){2-5}
Task &          0.1 &          0.3 &          1.0 &          3.0 \\
\midrule
       list 1st elem. &  0.00 (0.00) &  0.00 (0.00) &  0.89 (0.10) &  1.00 (0.00) \\
       list 2nd elem. &  0.00 (0.00) &  0.60 (0.15) &  0.70 (0.14) &  1.00 (0.00) \\
       list 3rd elem. &  0.00 (0.00) &  0.60 (0.15) &  0.60 (0.15) &  1.00 (0.00) \\
         list reverse &  0.00 (0.00) &  0.00 (0.00) &  0.70 (0.14) &  1.00 (0.00) \\
    list repeat twice &  0.00 (0.00) &  0.00 (0.00) &  0.50 (0.25) &  0.75 (0.22) \\
       list alt. even &  0.00 (0.00) &  0.00 (0.00) &  0.50 (0.20) &  1.00 (0.00) \\
        list alt. odd &  0.00 (0.00) &  0.00 (0.00) &  0.71 (0.17) &  0.86 (0.13) \\
 list circ. shift fw. &  0.00 (0.00) &  0.00 (0.00) &  0.00 (0.00) &  0.56 (0.17) \\
 list circ. shift bw. &  0.00 (0.00) &  0.00 (0.00) &  0.14 (0.13) &  1.00 (0.00) \\
      matrix diagonal &  0.00 (0.00) &  0.00 (0.00) &  0.50 (0.18) &  1.00 (0.00) \\
     matrix transpose &  0.00 (0.00) &  0.00 (0.00) &  0.50 (0.20) &  1.00 (0.00) \\
  matrix roll columns &  0.00 (0.00) &  0.00 (0.00) &  0.00 (0.00) &  0.17 (0.15) \\
 matrix elem. at idx. &  0.00 (0.00) &  0.50 (0.18) &  1.00 (0.00) &  1.00 (0.00) \\
\bottomrule
\end{tabular}}
\caption{Average in-context accuracy by task---with standard errors---as a function of CSCG overallocation on instruction-based prompts.}
\label{table:appendix_lialt_bytask_instructions_em_conv}
\end{table*}
\begin{table*}[htb!]
\footnotesize
\centering
\resizebox{.8 \textwidth}{!}{
\begin{tabular}{p{0.3\textwidth}P{0.15\textwidth}P{0.15\textwidth}P{0.15\textwidth}P{0.15\textwidth}}
\toprule
& \multicolumn{4}{c}{Overallocation ratio} \\
 \cmidrule(lr){2-5}
Task &          $0.1$ &          $0.3$ &          $1.0$ &          $3.0$ \\
\midrule
       list 1st elem. &  0.00 (0.00) &  0.00 (0.00) &  1.00 (0.00) &  1.00 (0.00) \\
       list 2nd elem. &  0.00 (0.00) &  0.12 (0.12) &  1.00 (0.00) &  1.00 (0.00) \\
       list 3rd elem. &  0.00 (0.00) &  0.38 (0.17) &  1.00 (0.00) &  1.00 (0.00) \\
         list reverse &  0.00 (0.00) &  0.00 (0.00) &  0.00 (0.00) &  0.88 (0.12) \\
    list repeat twice &  0.00 (0.00) &  0.00 (0.00) &  0.50 (0.18) &  0.88 (0.12) \\
       list alt. even &  0.00 (0.00) &  0.00 (0.00) &  0.00 (0.00) &  1.00 (0.00) \\
        list alt. odd &  0.00 (0.00) &  0.00 (0.00) &  0.50 (0.18) &  1.00 (0.00) \\
 list circ. shift fw. &  0.00 (0.00) &  0.00 (0.00) &  0.00 (0.00) &  0.88 (0.12) \\
 list circ. shift bw. &  0.00 (0.00) &  0.00 (0.00) &  0.29 (0.17) &  1.00 (0.00) \\
      matrix diagonal &  0.00 (0.00) &  0.67 (0.19) &  0.67 (0.19) &  1.00 (0.00) \\
     matrix transpose &  0.00 (0.00) &  0.00 (0.00) &  0.43 (0.19) &  1.00 (0.00) \\
  matrix roll columns &  0.00 (0.00) &  0.00 (0.00) &  0.00 (0.00) &  0.50 (0.18) \\
 matrix elem. at idx. &  0.00 (0.00) &  0.38 (0.17) &  1.00 (0.00) &  1.00 (0.00) \\
\bottomrule
\end{tabular}}
\caption{Average in-context accuracy by task---with standard errors---as a function of CSCG overallocation on example-based prompts.}
\label{table:appendix_lialt_bytask_examples_em_conv}
\end{table*}

\newpage
\subsection{Example failures}
Finally, we present a few examples which illustrate the failure modes of our approach. These are primarily a consequence of imperfections in the learned CSCG model.

Each example is presented in the format (prompt, \textcolor{OliveGreen}{ground truth correct output}, \textcolor{red}{actual model response}). 

\begin{enumerate}
    \item For these failures, the instruction circuit has been wired to the wrong algorithm circuit (possibly driven by the ambiguity of the forward slash delimiter separating the instruction from the example), resulting in the retrieval of the wrong schema.
    {\footnotesize
    \begin{itemize}
        \item \texttt{output odd indexed elements / [ U V B Q K I ]	} \\ \textcolor{OliveGreen}{\texttt{[ U B K ] /}} \\ \textcolor{red}{\texttt{[ V Q I ] /	}}
        \item \texttt{flip the list / [ S E J ]} \\ \textcolor{OliveGreen}{\texttt{[ J E S ] /}} \\ \textcolor{red}{\texttt{[ S S E E J J ]}}
        \item \texttt{reverse the list / [ R T B ]} \\ \textcolor{OliveGreen}{\texttt{[ B T R ] /}} \\ \textcolor{red}{\texttt{[ R R T T B B ] /}}
        \item \texttt{mirror the list / [ B A O T ]} \\ \textcolor{OliveGreen}{\texttt{[ T O A B ] /}} \\ \textcolor{red}{\texttt{[ B B A A O O T T ] /}}
    \end{itemize}}

    \item For these failures, the schema has been learned incorrectly.
    {\footnotesize
    \begin{itemize}
        \item \texttt{switch the items of the list one position forward / [ L N G X M T ]} \\
        \textcolor{OliveGreen}{\texttt{[ T L N G X M ] /}} \\ \textcolor{red}{\texttt{[ T L N G X M T ] [ T L N G X M T ] $\ldots$}}
        \item \texttt{shift the columns of the matrix to the right / [ [ D Y ] [ V F ] ]} \\ \textcolor{OliveGreen}{\texttt{[ [ Y D ] [ F V ] ] /}} \\ \textcolor{red}{\texttt{[ [ get }}
        \item \texttt{/ [ Z J B ] [ Z Z J J B B ] / [ B A E F W L ]} \\ \textcolor{OliveGreen}{\texttt{[ B B A A E E F F W W L L ] /}} \\ \textcolor{red}{\texttt{[ B B A E F F W W L L ] /}}
        \item \texttt{/ [ V P X T ] [ P T ] / [ V F J P E W ]} \\ \textcolor{OliveGreen}{\texttt{[ F P W ] /}} \\
        \textcolor{red}{\texttt{[ F P W ] [ F P W ] $\ldots$}}   
    \end{itemize}}
\end{enumerate}

\end{document}